# Markov Chain Neural Networks

Maren Awiszus, Bodo Rosenhahn
Institut für Informationsverarbeitung
Leibniz Universität Hannover
www.tnt.uni-hannover.de

## Abstract

*In this work we present a modified neural network model which is capable to simulate Markov Chains. We show how to express and train such a network, how to ensure given statistical properties reflected in the training data and we demonstrate several applications where the network produces non-deterministic outcomes. One example is a random walker model, e.g. useful for simulation of Brownian motions or a natural Tic-Tac-Toe network which ensures non-deterministic game behavior.*

## 1. Introduction

A Markov model is a mathematical model to represent a randomly changing system under the assumption that future states only depend on the current state (Markov property). It is used for predictive modeling or probabilistic forecasting. A simple model is a Markov chain which models a path through a graph across states (vertices) with given transition probabilities on the edges. An example application is a random walker generated by sampling from a joint distribution using Markov Chain Monte Carlo. Random walks have applications in various fields, such as physics, ecology, economics, chemistry or biology as they serve as a fundamental model to record stochastic activity. Besides Random walker models, various computer science applications require controlled stochastic behavior, e.g. to generate AIs with smart but non-deterministic patterns for game play or for a more natural behavior of Chat Bots in Human-Computer Interaction.

Artificial Neural networks are the current state-of-the art method for solving various computer vision and machine learning tasks. Due to the increased computational possibilities using GPUs and the availability of big data, top rank scores in various challenges are obtained with deep neural networks. Due to their underlying concept of connected neurons across hidden layers, once trained, a neural network behaves in a deterministic fashion. Taking an interactive game or a human-computer chat as examples, it leads (a) to a foreseeable reaction given a specific game configuration or (b) always to the same answer for a given comment in a dialog system. Overall, it results in a non-natural behavior for human-computer interaction. Indeed, when training a neural network with different possible (but legal) outcomes, it leads to slow convergence but still deterministic behavior when e.g. using a soft-max decision. A common solution to this issue is to train in a network the distributions of possible outcomes and then to sample from this. Probabilistic sampling from a regression network requires a post-processing step in the test phase which is undesired in many applications.

**Contributions**

In this paper we address all above aspects and present a neural network model which is capable to (a) simulate Markov Chains, (b) we show how to train such a network with given statistical properties reflected in the training data and (c) demonstrate several applications where the network produces random outcomes. Since the stochastic decision process is integrated by using a random node value as additional input, no post-processing (e.g. sampling from a result-distribution) is necessary.

Experiments are conducted for (a) a probabilistic graphical model (e.g. a 2D and 3D random walker) (b) a natural Tic-Tac-Toe or Flappy Bird gameplay (c) a text-synthesizer given an input text database (in our example from a poem or books like *Moby Dick* from Herman Melville or *Curious George* by H.A. and M. Rey and (d) for image completion (MNIST) or the synthesis of facial emotions.

## 2. Foundations and Stochastic Neural Networks

This section summarizes the foundations and some state-of-the art for this work with a focus on Markov Chains and Neural Networks.

### 2.1. Graphical Models

A probabilistic graphical model (PGM) is a probabilistic model for which a graph $G = (V, E)$ with given edges

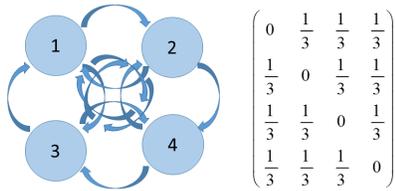

Figure 1. Example Markov Model

$E$ and vertices $V$ is used to model the conditional dependence between random variables [12]. Bayesian networks or Markov random fields are famous examples with numerous successful applications in computer vision and machine learning [2, 6, 28, 18]. A stochastic process has the so-called Markov property if the conditional probability distribution of future states of the stochastic process only depends on the current state and not on the sequence of events that preceded it. Thus for $S$ being a discrete set with a respective discrete sigma algebra the Markov property is given as

$$P(X_n = x_n | X_{n-1} = x_{n-1}, \ldots, X_1 = x_1) = P(X_n = x_n | X_{n-1} = x_{n-1})$$

Well known are two types of Markov models, (a) the visible Markov model, and (b) the hidden Markov model or HMM. In (visible) Markov models (like a Markov chain), the state is directly visible to the observer, and therefore the state transition (and sometimes the entrance) probabilities are the only parameters, while in the hidden Markov model, the state is hidden and the (visible) output depends on the (non-visible) state. The most likely hidden states can be recovered e.g. using the famous Viterbi algorithm [27]. Thus, each state has a probability distribution over the possible output tokens. In this work we will focus on Markov Chains, simply given as a Graph $G = (V, E, T)$ with $E \subseteq V \times V$ and $T_{i,j} = p(i|j)$ for $i, j \in E$. Where $T$ gives the transition probabilities along the edges between vertices. Random walks (as considered later) or the *Gambler's ruin problem* are famous examples of Markov Chain processes. For further details, the reference [1] is highly recommended.

## 2.2. Neural Networks

A neural network is a collection of connected neurons [10]. Each (artificial) neuron is defined as a weighted sum of input values (given as inner product and an added bias value) passed on to a so-called activation function (e.g. a sigmoid function or a linear function [19]) to produce an output (or activation). Combining an input vector with several neurons yields a so-called layer and connecting the outcome of a layer to further layers leads to a neural network. Such a network is usually optimized by gradient descent on an output error using backpropagation. In this work we used the well established stochastic gradient descent method [3]. Due to the huge amount of parameters to optimize, weight sharing (using e.g. convolutions [9, 20]) or local receptive fields [26] can drastically avoid overfitting and support the convergence of the neural network. In this paper we do not discuss the different variants of neural networks and their possibilities for optimization, autoencoders [17], incremental learning [14, 15] or data management [4]. We only want to state, that neural networks are commonly used for competing in different benchmarks with remarkable performance [13, 21, 11]. For this work it is only important to clarify that a neural network is a deterministic function and in its nature not suited for modeling Markov chains.

## 2.3. Stochastic and probabilistic neural networks

The most similar approach to our proposed contribution are so-called stochastic neural networks [25]. They are a type of artificial neural networks where random variations are built into the network. This can be realized by modifying the involved neurons with stochastic transfer functions, or by giving them stochastic weights. These modifications can be useful for optimization problems, since the random fluctuations can help to overcome local minima [8]. In a similar fashion, noisy activation functions have been proposed [7]. To our experience, the stochastic weights or activations can lead to unpredictable network behavior, since a slight change in an early layer can propagate through the whole network. Therefore, it is nearly impossible to ensure desired statistical properties of such networks, unless they are explicitly modeled [24]. A probabilistic neural network (PNN) is a special feed-forward neural network [22, 5]. Herefore, the probability distribution of each possible class is approximated by a distribution, e.g. using a Parzen window and a non-parametric function. Then for a new input vector the PDF for each possible class is evaluated and Bayes rule is applied for the final decision. The allocated class is the class of the highest posterior probability. The resulting network consists of four layers, namely the input layer, hidden layer, pattern layer and the output layer. Since the pattern layer is usually representing the likelihood for a specific class, sampling from this layer could be done to produce random outcomes. This again leads to an extra analysis step during testing which we will avoid in this paper. Instead, we shift the information of the outcome distribution into the training phase to sample solutions directly from the network. This will be achieved by an extra random input value for the first layer.

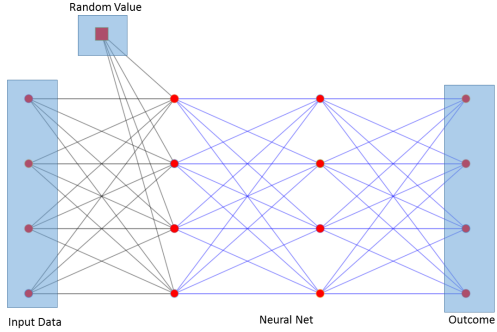

Figure 2. Markov Chain Neural Network

## 3. Markov Chain Neural Network

In the following we describe the basic idea for our proposed non-deterministic MC neural network, suitable to simulate transitions in graphical models. Similar to the previous section we start with a Graph $G = (V, E, T)$ with $V$ a set of states, $E \subseteq V \times V$ and a matrix with transition probabilities $T = V \times V$ with $\sum_i^{|V|} T(i,j) = 1$. A simple example is shown in Figure 1 with the states $1 \ldots 4$ and the respective transition matrix.

This simple example can be used to model e.g. a random walker with 4 oriented steps (*left*, *right*, *up*, *down*) with the property that the walker is changing the orientation with every step. Given a graphical model and an initial state (e.g. $\mathbf{1} = (1, 0, 0, 0)^T$ representing the first node the next states are $2, 3, 4$ with a likelihood of $\frac{1}{3}$. An obviously naive option to train a network is to define a four-dimensional (binary) input and output vector, to generate training examples while traversing through the network and to use this to train a neural net. Unfortunately, a neural net behaves in a deterministic way, so that for a given input, the outcome of a network is always the same. Alternatively, the transition probability of $[0, \frac{1}{3}, \frac{1}{3}, \frac{1}{3}]^T$ can be trained and then with a sampling on the target distribution, a random decision can be made. To allow for non-deterministic behavior and to avoid the sampling from a target distribution, the goal is to transfer the predefined statistical behavior of a graphical model to a neural net. To achieve this, we propose the extension of the input data with an additional value containing a random number $r \in [0,1]$ and a special learning paradigm, described in the following.

The topology is visualized in Fig. 2. The random number is connected to the neural net as additional input value during training and testing. It can be implemented by simply using a five-dimensional input vector $(r, 1, 0, 0, 0)$ following the above example, or by using an additional random bias value, with a random value for a test input. The key idea is that the random value steers the output vector following the predefined statistical behavior: For a given training set $\mathcal{D}$ with input vector $x_i$ and output vector (or value) $y_i$,

$$\mathcal{D} = \{(x_i, y_i)\}_{i=1}^n$$

we approximate $p(y_i | X = x_i)$ as a (in general multivariate) discrete probability distribution. Thus for $y_i \in 1 \ldots c$, and $Y = \{y_1 \ldots y_c\}$ and a given input vector $X$,

$$\sum_{i=1}^c p(y_i | X) = 1$$

E.g. assuming the transition probabilities from Figure 1, it implies the row-sum to be one. From the training data it is simply approximated as the relative frequency (the empirical probability),

$$p(y_i | x_i) = \frac{\| \{(x = x_i, y = y_i) \in \mathcal{D}\} \|}{\| \{(x = x_i, y \in Y) \in \mathcal{D}\} \|}$$

Now, it is possible to generate from such a histogram an arbitrary number of input/output pairs. They reflect a predefined reaction, given a random value as additional input node so that the distribution of possible output vectors corresponds to the distribution of output vectors in the training data. Thus, we augment the input data vectors with an additional random value and use these random values to draw a distribution of possible outputs.

Similar to importance sampling [23], we accumulate for each possible input state the corresponding cumulative frequency, e.g. for our simple random walker model and start node $(1, 0, 0, 0)^T$ we gain $[0, \frac{1}{3}, \frac{2}{3}, 1]$. Then we generate new training data by drawing a random number $r$ and by identifying the appropriate decision from the accumulated interval. The following example shows some example training data which are generated from this strategy:

| Input | Output |
|---|---|
| $(0.5, 1, 0, 0, 0)$ → | $(0, 0, 1, 0)$ |
| $(0.2, 1, 0, 0, 0)$ → | $(0, 1, 0, 0)$ |
| $(0.8, 1, 0, 0, 0)$ → | $(0, 0, 0, 1)$ |
| $(0.9, 1, 0, 0, 0)$ → | $(0, 0, 0, 1)$ |
| $(0.1, 1, 0, 0, 0)$ → | $(0, 1, 0, 0)$ |
| $\ldots$ | |

Figure 3 shows some example random walks which have been generated from a neural net by feeding the output as new input vector into the net. The images show the realization of 20.000 steps with $(1, 0, 0, 0)$ as start node. The state probability converges to $\frac{1}{4}$, thus each state has the same visiting expectation likelihood. Simply speaking, the additional random value acts as a switch which defines in a predefined manner the respective outcome. E.g. for a start node $(1, 0, 0, 0)$ and a random value between $[0, \frac{1}{3}]$ the outcome is $(0, 1, 0, 0)$, whereas for a random value between $]\frac{1}{3}, \frac{2}{3}]$ the outcome is $(0, 0, 1, 0)$, etc. Figure 4 shows examples of a neural net which generates 3D random walks.

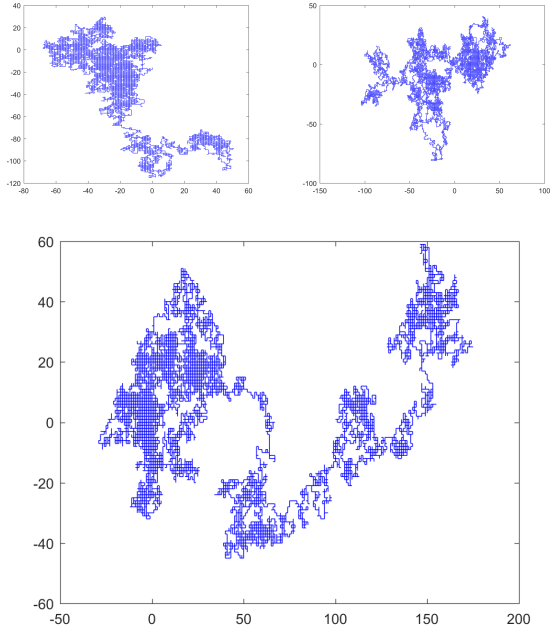

Figure 3. Realizations of different 2D random walks generated with a Markov Chain Neural Network

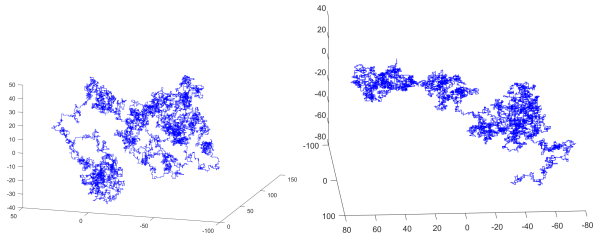

Figure 4. Realizations of different 3D random walks generated with a Markov Chain Neural Network

## 4. Experiments

Starting from this toy example, we now demonstrate further examples on how to use the MC-neural network.

### 4.1. Tic-Tac-Toe

The famous game is a paper-and-pencil game for two players, X (*black*) and O (*white*). They take turns by marking the spaces in a $3 \times 3$ grid. The player who succeeds in placing three own marks in a horizontal, vertical, or diagonal row wins the game. Due to its simplicity, there are only 26,830 possible games up to rotations and reflections, thus it is a nice example for a neural network to learn.

For the generation of training data we implemented a *non-stupid* rule-based player which follows the following steps

1. Can I win ?

2. Do I have to defend ?

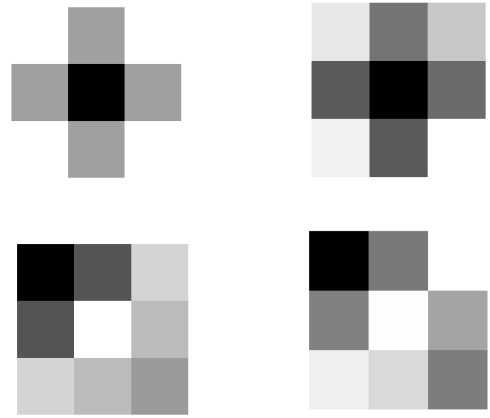

Figure 5. Distribution of possible reactions during a Tic-Tac-Toe game. A high gray value (towards white) indicates a high probability whereas a low gray value (towards black) indicates a low probability. The selected field has the likelihood 0 since it can not be selected anymore. Left : Ground truth (the statistics of the training data reactions when (top) the middle place has been selected or (b) when the upper left corner has been selected. Right: Distribution of reactions of our trained MC network during 1000 artificial games with the selected middle or upper left space as starting point. As can be seen, the neural network produces reactions which are close to the distribution of reactions in the training data.

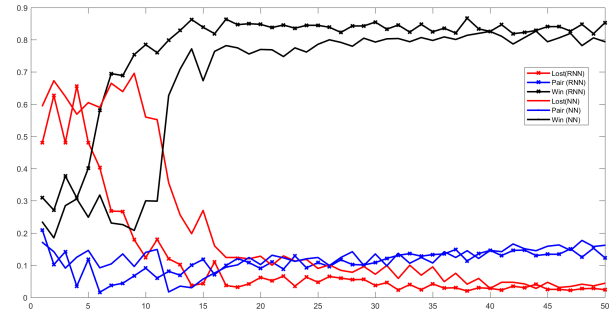

Figure 6. Performance over several Epochs of training. After each training epoch, 1000 games between the MC neural network (lines with dots) and the rule-based player (solid lines) are simulated. It results in a ratio of win (black), pair (blue) and lost (red) games. The diagram shows, that the Markov Chain neural network converges faster and yields a slightly better playing performance. (The Figure is best viewed in color)

3. Can I make a move to build a fork ?

4. Make a random move

We simulated several thousand games and use the reactions of the winner during the games as output and as input the configuration of the preceding step. Thus, the input and output configuration is a nine-dimensional vector with values $[-1, 0, 1]$. The value $-1$ represents the black player $X$, $0$ is an empty field and $1$ is the white player $O$. As neural network we decided for a simple shallow fully connected structure with $[10 : 80 : 30 : 9]$ layers, where the fist layer of dimension 10 is the configuration and the additional random value and 9 is the outcome configuration.

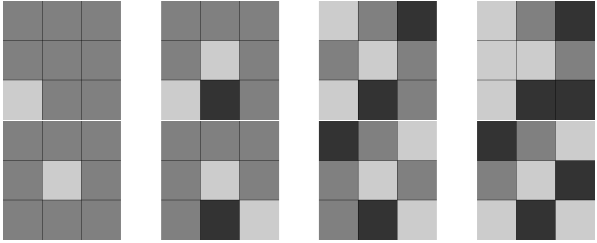

Figure 7. Two example games of our proposed neural network. The neural network (white) starts, produces different start configurations at the first move and is later able to generate a fork scenario so that the black player (the rule based player) looses the game.

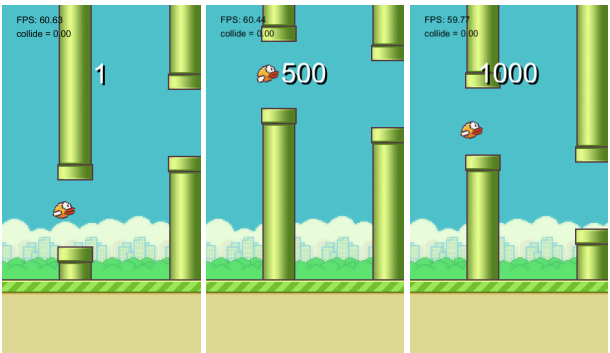

Figure 8. A neural network playing *Flappy Bird* perfectly.

From the games, it is now easy to determine from the training data the amount of possible reactions which have some usefulness for the neural net. E.g. when considering Figure 5, the upper left image shows the distribution of possible reactions to a start configuration of player X(*black*) who has selected the middle field. Thus it can be seen, that the best option is to react by using a corner field (bright value). In the lower left example, player X(*black*) selected the upper left corner and the worst reaction is to select a direct neighbor (right, bottom) field. Indeed, if a user selects one of these fields, there is a $100\%$ winning strategy for the *black* player X.

If player X(*black*) starts with a middle field, due to the symmetry properties it is not important with which corner player O (*white*) reacts, so that all remaining fields have a certain likelihood for a reaction, which can be estimated from the training data and embedded into the MC neural net, as described in the previous section. The right image of Figure 5 shows the distribution of reactions after 1000 trials with a predefined start configuration which is the X (*black*) player selecting the middle field (top) or upper left field (bottom). It is clearly visible, that the distribution is very close to the distribution of the training data. Thus, a useful natural reaction pattern is trained and the network can play in a non-deterministic but appropriate fashion. We tested the neural network with several players and all confirmed the natural gameplay of our network. One reason is also, that there is a probability for the network to produce non-optimal and variable moves.

Figure 6 demonstrates the performance of the trained MC neural network over training several epochs in comparison to a classical network. The MC network converges faster. In the authors opinion one reason is, that the same input in the training data can lead to different outputs. Thus, the gradients can start to contradict each other yielding to a slower convergence. For the MC network, the additional random number allows for a suited separation of the behaviors and thus a faster and better convergence and game play.

Figure 7 shows two games between the Markov Chain neural network and the rule based player. The neural network starts (white) and already as first move, different configurations are produced by the network. In the remainder of the game, the network produces a fork scenario, so that the rule based player looses the game.

### 4.2. Reinforcement Learning

This paradigm can also be applied in the context of reinforcement learning to balance possible reactions to their overall gain. In Q-learning an agent transitions between states, executes actions and gains a reward to be optimized. The non deterministic behavior of a Markov chain neural network can be easily integrated in an agent to explore the state space of a game. The rewards are correlated to the impact of an action, so that more successful activities appear more often in the training data and are thus more likely to be selected. Figure 8 shows three screen shots of a neural network which perfectly plays the game *Flappy Bird*, in which the inputs consist of the proposed random value, the position of the bird, the position of the pipes and the distance between the bird and the pipes.

### 4.3. Text synthesis

The next example is text synthesis. Based on given input letters (we use 7 letters as input, which are encoded as their ASCII-value), the network predicts a new letter to continue the text. This allows the synthesis of new text blocks, e.g. useful for artificial chat-bots. In this experiment we use the poem *The cat with the hat* by Dr. Seuss. It has a length of 1620 words and 7086 characters. Given 7 input characters, it is now possible to determine the statistics for the follow up character and to train the Markov Chain neural network. Whereas the input is an 8-dimensional vector containing the Markov Chain and the ASCII-values of the characters, the output is a 256 dimensional decision vector with binary values, indicating the decision as ASCII value.

After training the network, it is possible to start with a fragment and then to continue the synthesis of the text. Figure 9 shows on the left an input text fragment of the network and in the right some synthesized sentences. The figure also indicates repetitions of characters in the text, so that it is easy to verify how the network *jumps* through the input text while synthesizing the text.

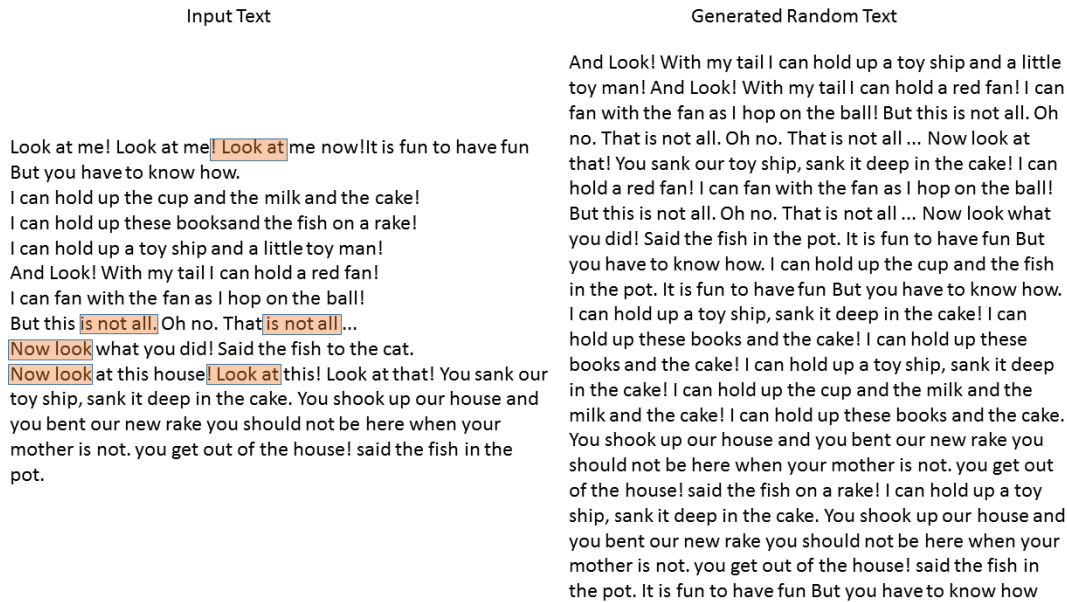

Figure 9. Left: Input Text, fragment from The cat with the hat (Dr. Seuss). Right: Synthesized text with Markov property along the words.

We also trained a text-synthesizer from books like *Moby Dick* or *Curious George*. Whereas the first contains 211.000 words which are arranged in a 211.000 dimensional dictionary vector, the second one only contains 950 words yielding a much smaller dictionary. Here the intention is not to learn successive letters, but consecutive words, directly from the local co-occurrence of the surrounding words. Thus, for training we look up for each position in the text a predefined amount of consecutive words (e.g. 6). E.g. from the sentence *There now is your insular city of the Manhattoes, belted round by wharves as Indian isles by coral reefs-commerce surrounds it with her surf.* We select *belted round by wharves as Indian* and lookup their respective dictionary entry numbers, e.g. $[120, 745, 823, 774890, 132]$. To account for the order of the words, the input training vector is zeros and the values of the above positions are initialized with $\frac{1}{7-i}$. Thus, the words are ordered with respect to the positions by using the weight in the input vector. Thus, words close to the *next* words have a higher weight and therefore a higher importance for prediction. For both example books, after training of a 5 layer neural network, the network was able to produce realistic text fragments. An example produced by the network is *The man took off the bag. George sat on a little boat, and a sailor rowed them both across the water to a big Zoo in a prison.*

Even though the generated text appears partially useful, the overall generated sentences are not very smart since the global context of the text is ignored.

### 4.4. Image completion

In the next example we use the MNIST database [16] and train a network which uses as input the upper left quarter of an image and as output the complete image. Thus, the goal of the network is to fill in the missing information in the image. As the solution to the input is ambiguous, a classical neural network has severe problems to find an appropriate mapping and thus, ends up in a mixture of possible solutions, see Figure 10. In contrast, our MC network allows for a clear separation of possible solutions, so that several possible answers are generated, see Figure 11.

For this experiment we used a simple shallow network with 500 hidden units.

In the following example we use the jaffee-database as data source. The database consists of several frontal face images of actors performing different emotions (anger, disgust, fear, happy, surprise, neutral).

Our application uses an image part as input and our MC neural network to generate a similar face with a different emotion. For this we first classify the ID of the person and use this input value, together with the random value as input for our network. If we use an equal distribution for the emotion changes, different random emotions are generated independently from the current emotion. Some examples are shown in the right of Figure 12.

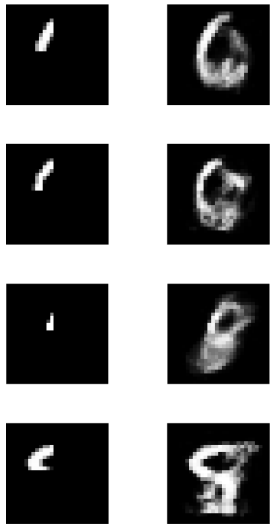

Figure 10. Left: Input Image fragment. Right: Outcome of a deterministic Neural network. As can be seen, mixtures of possible solutions are generated.

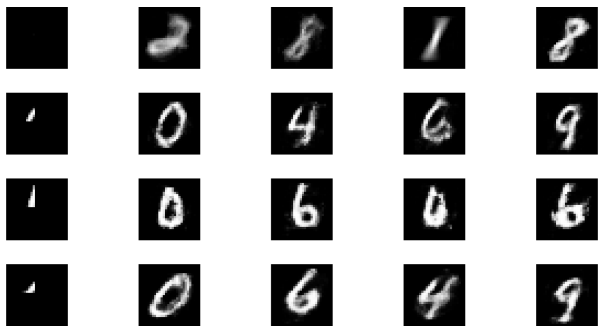

Figure 11. Left: Input Image fragment. Right: Different generated solutions from the proposed Markov Chain network.

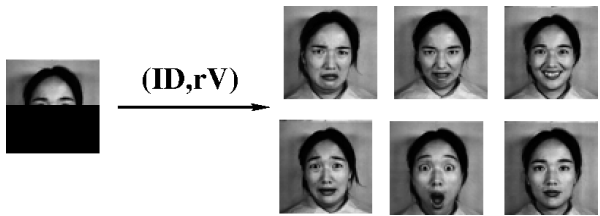

Figure 12. Left: Input Face part (used for identification). Right: Synthesized faces with Markov property along emotions.

## 5. Summary and Discussion

In this work we present a modified neural network model which is capable to simulate Markov Chains. We show how to train such a network with given statistical properties reflected in the training data and demonstrate several applications where the network produces random outcomes for generating a random walker model or a natural Tic-Tac-Toe gameplay. The key idea is to add an additional input node with a random variable which allows the network to use it as a switch node to produce different outcomes. Even though the network is acting in a deterministic fashion, due to the random input it produces random output with guaranteed statistical properties reflected in the training data. The MC network is based on a statistical analysis of the training data and does not require further post-processing (e.g. sampling from a distribution of solutions). The network is straight forward to implement. It allows natural game play, ambiguous image completion or a more natural chat avatar as possible application.